\pdfoutput=1

\documentclass[11pt]{article}

\usepackage{ACL2023}
\usepackage{times}
\usepackage{latexsym}

\usepackage[T1]{fontenc}

\usepackage[utf8]{inputenc}

\usepackage{microtype}
\usepackage{times}
\usepackage{latexsym}
\usepackage{soul}
\usepackage{amsmath}
\usepackage{booktabs}
\usepackage{bm}
\usepackage{cleveref}
\usepackage{graphicx}
\usepackage{tabularx}
\usepackage{soul}
\usepackage{xcolor}
\usepackage{todonotes}
\usepackage{multirow}
\usepackage{xpatch}
\usepackage{blindtext}
\usepackage{fdsymbol}
\usepackage{microtype}
\usepackage{hyperref}
\usepackage{adjustbox}
\usepackage{textcomp}
\usepackage{xparse}
\usepackage{enumitem}
\usepackage{makecell}
\usepackage{subcaption}
\usepackage{colortbl}

\usepackage{xparse}

\crefformat{section}{\S#2#1#3} %
\crefformat{subsection}{\S#2#1#3}
\crefformat{subsubsection}{\S#2#1#3}

\DeclareMathOperator*{\argmax}{argmax}   %

\newcommand{\samsum}[1]{\textsc{SAMSum}}
\newcommand{\dialogsum}[1]{\textsc{DialogSum}}
\newcommand{\mixandmatch}[1]{\textsc{MixAndMatch}}
\newcommand{\confit}[1]{\textsc{ConFiT}}
\newcommand{\ctrldiasumm}[1]{\textsc{CtrlDiaSumm}}
\newcommand{\cods}[1]{\textsc{CODS}}
\newcommand{\modelshort}[1]{\textsc{ZeroFEC}}
\newcommand{\modelshortda}[1]{\textsc{ZeroFEC-DA}}
\newcommand{\pgn}[1]{\textsc{PGN}}
\newcommand{\cliff}[1]{\textsc{CLIFF}}
\newcommand{\conseq}[1]{\textsc{ConSeq}}

\definecolor{c2}{RGB}{218,0,0}

\definecolor{lightblue}{RGB}{212, 235, 255}
\definecolor{lightorange}{RGB}{255, 204, 168}
\definecolor{lightyellow}{RGB}{255, 255, 168}
\definecolor{lightred}{RGB}{255, 168, 168}
\definecolor{darkred}{RGB}{196, 30, 58}
\definecolor{lightgreen}{rgb}{0.85, 0.85, 0.85}
\definecolor{gold}{rgb}{0.83, 0.69, 0.22}
\sethlcolor{lightblue}
\newcolumntype{Y}{>{\centering\arraybackslash}X}
\newcommand\hlc[2]{\sethlcolor{#1} \hl{#2}}
\definecolor{gold}{rgb}{0.83, 0.69, 0.22}

\newcommand{\markred}[1]{{\color{darkred}#1}}

\title{Zero-shot Faithful Factual Error Correction}

\author{Kung-Hsiang Huang\textsuperscript{$\spadesuit$}~~~ Hou Pong Chan\textsuperscript{$\heartsuit$} ~~~ Heng Ji\textsuperscript{$\spadesuit$}\\
\textsuperscript{$\spadesuit$}Department of Computer Science, University of Illinois Urbana-Champaign \\
\textsuperscript{$\heartsuit$}Faculty of Science and Technology, University of Macau \\
\textsuperscript{$\spadesuit$}\texttt{\{khhuang3, hengji\}@illinois.edu} \\
\textsuperscript{$\heartsuit$}\texttt{hpchan@um.edu.mo}
  \\}

\begin{document}
\maketitle

\begin{abstract}
Faithfully correcting factual errors is critical for maintaining the integrity of textual knowledge bases and preventing hallucinations in generative models. Drawing on humans' ability to identify and correct factual errors, we present a zero-shot framework that formulates questions about input claims, looks for correct answers in the given evidence, and assesses the faithfulness of each correction based on its consistency with the evidence. Our zero-shot framework outperforms fully-supervised approaches, as demonstrated by experiments on the \textsc{FEVER} and \textsc{SciFact} datasets, where our outputs are shown to be more faithful. More importantly, the decomposability nature of our framework inherently provides interpretability. Additionally, to reveal the most suitable metrics for evaluating factual error corrections, we analyze the correlation between commonly used metrics with human judgments in terms of three different dimensions regarding intelligibility and faithfulness.\footnote{The code and data have been made publicly available: \url{https://github.com/khuangaf/ZeroFEC}} \looseness=-1
\end{abstract}
\section{Introduction}

The task of correcting factual errors is in high demand and requires a significant amount of human effort. The English Wikipedia serves as a notable case in point. It is continually updated by over 120K editors, with an average of around six factual edits made per minute\footnote{\url{https://en.wikipedia.org/wiki/Wikipedia:Statistics}}. Using machines to correct factual errors could allow the articles to be updated with the most current information automatically. This process, due to its high speed, can help retain the integrity of the content and prevent the spread of false or misleading information. 

In addition, the hallucination issues have been shown to be a prime concern for neural models, where they are prone to generate content factually inconsistent with the input sources due to the unfaithful training samples \cite{maynez-etal-2020-faithfulness} and the implicit ``knowledge'' it learned during pre-training \cite{niven-kao-2019-probing}. Factual error correction can be used in both pre-processing and post-processing steps to rectify the factual inconsistencies in training data and generated texts, respectively. This can help build trust and confidence in the reliability of language models.

\begin{figure}[bt]
    \centering
    \includegraphics[width=0.9\linewidth]{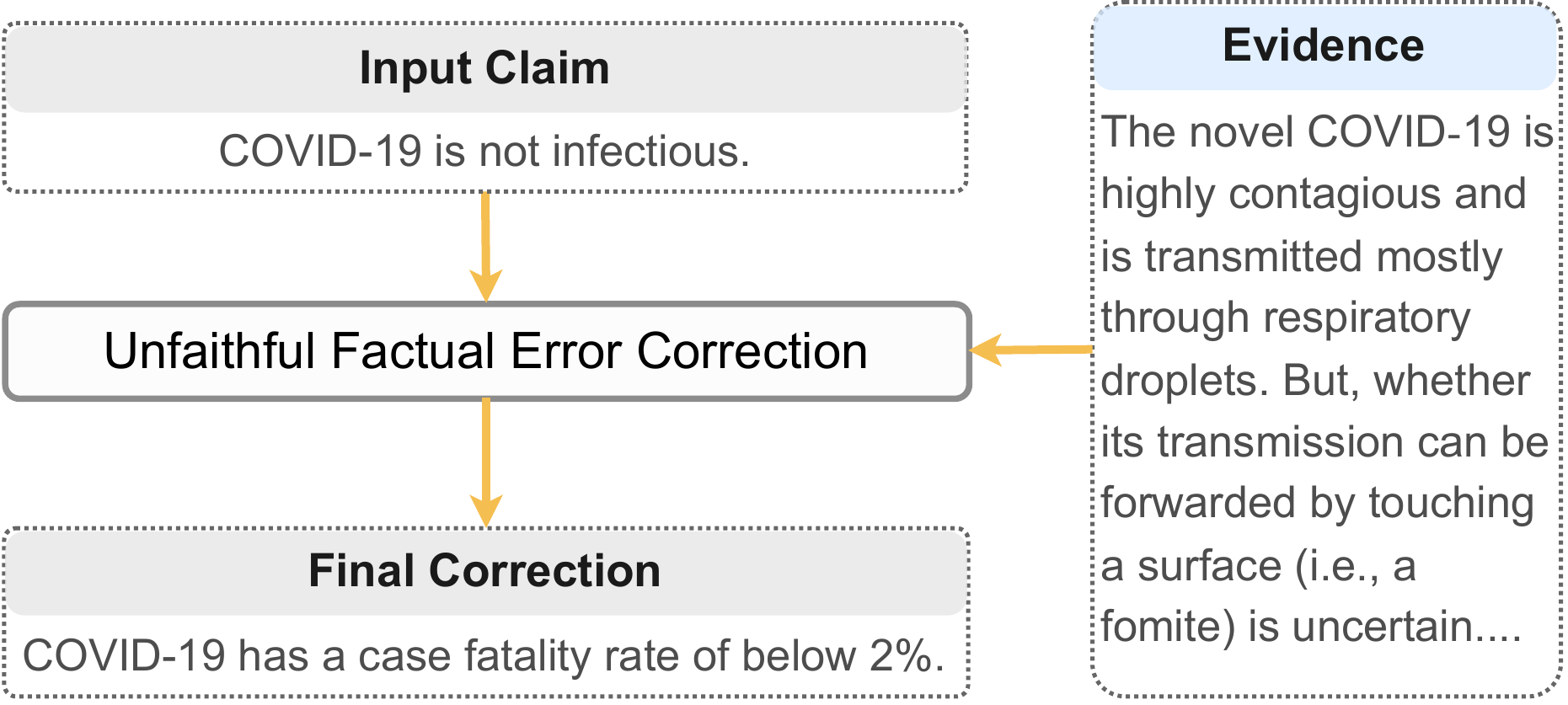}
    \caption{An example of a factual but unfaithful correction leading to misleading information. While it is technically true that the majority of people infected with COVID-19 will recover, there is no information in the evidence that supports the final correction. Additionally, when this statement is taken out of context, it could mislead people to believe that COVID-19 is not dangerous and that there is no need for precautions, which is false. A factual and faithful correction is ``COVID-19 is highly contagious.''.\looseness=-1}%
    \label{fig:toy_example}
\end{figure}

Prior work %
typically formulates factual error correction as a sequence-to-sequence task, either in a fully supervised or in a distantly supervised manner \cite{shah2020automatic, thorne-vlachos-2021-evidence}. While these approaches have made great strides in generating fluent and grammatically valid corrections, they only focus on the aspect of \textit{factuality}: \textit{whether the outputs are aligned with facts}. Little emphasis was placed on \textit{faithfulness}: \textit{the factual consistency of the outputs with the evidence}. Faithfulness is critical in this task as it indicates whether a generated correction reflects the information we intend to update. If faithfulness is not ensured, this could lead to the spread of misleading content, causing serious consequences. \Cref{fig:toy_example} shows a concrete example. In the context of automatically updating textual knowledge bases, the topic of an unfaithful output would likely deviate much from that of the expected correction. Therefore, such an edit is not desirable, even if it is factual.

\begin{figure*}[t]
    \centering
    \includegraphics[width=0.9\linewidth]{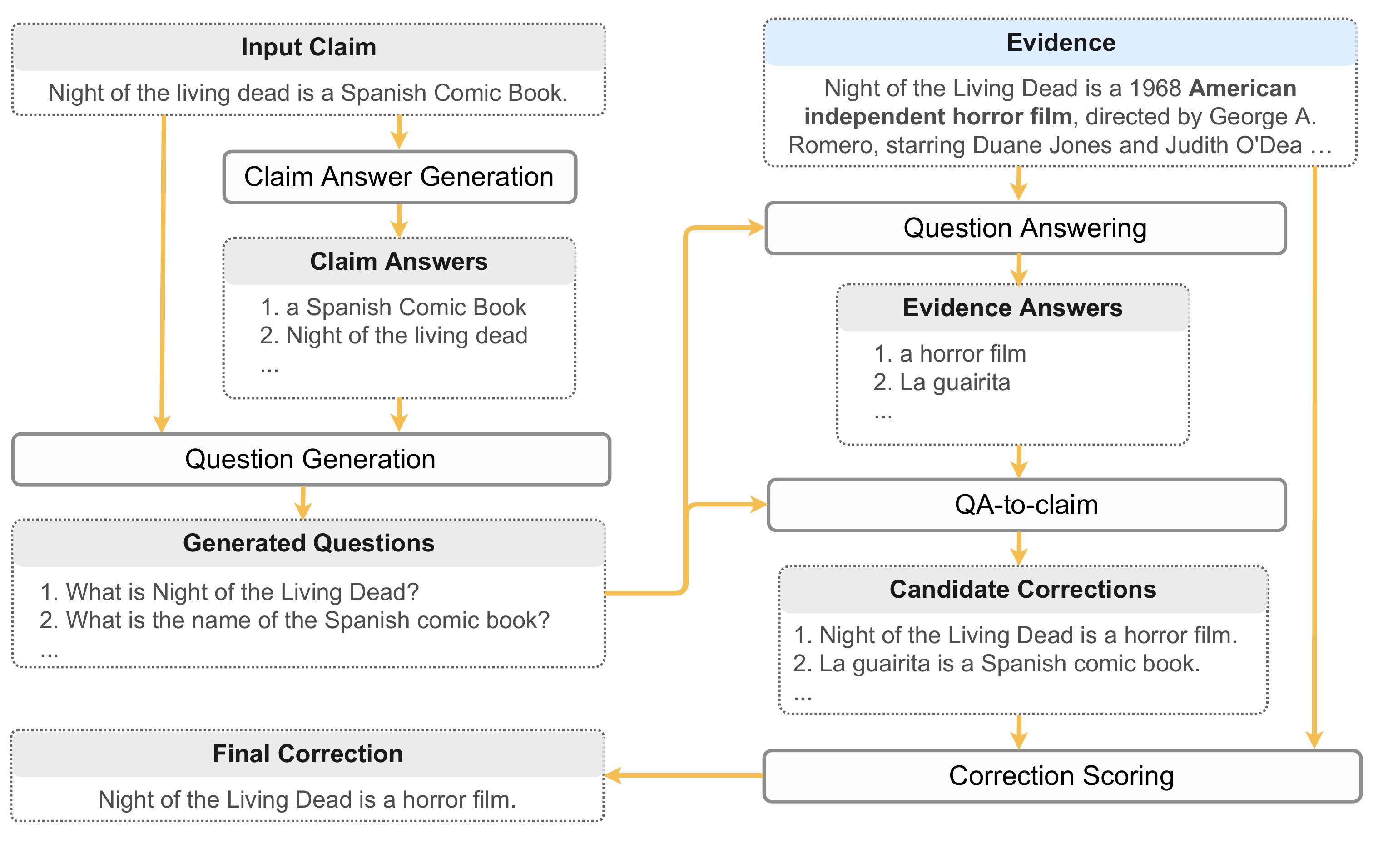}
    \caption{An overview of our framework. First, given an input claim, we generate the \textit{claim answers} by enumerating all information units in the input claim. Second, conditioned on each extracted answer and the input claim, a question is generated. Third, each question is then fed to a question answering model to produce an \textit{evidence answer} using the given evidence as context. Fourth, using a sequence-to-sequence approach, each \textit{evidence answer} and the corresponding question are transformed into a statement, %
    which serves as a \textit{candidate correction}. Finally, the \textit{final correction} is produced by scoring candidate corrections based on faithfulness.} %
    \label{fig:framework_overview}
\end{figure*}

In this work, we present the first study on %
the \textit{faithfulness} aspect of factual error correction. To address faithfulness, we propose a \textit{zero-shot} factual error correction framework (\modelshort~), inspired by how humans verify and correct factual errors. When humans find a piece of information suspicious, they tend to first identify potentially false information units, such as noun phrases, then ask questions about each information unit, and finally look for the correct answers in trustworthy evidence \cite{saeed2022crowdsourced, chen2022generating}.
Following a similar procedure, \modelshort~ breaks the factual error correction task into five sub-tasks: %
(1) \textit{claim answer generation}: extracting all information units, such as noun phrases and verb phrases, from the input claim; %
(2) \textit{question generation}: generating question given each \textit{claim answer} and the original claim such that each \textit{claim answer} is the answer to each generated question; %
(3) \textit{question answering}: answering each generated question using the evidence as context; (4) \textit{QA-to-claim}: converting each pair of generated question and answer to a declarative statement; (5) \textit{correction scoring}: evaluating corrections based on their faithfulness to the evidence, where faithfulness is approximated by the entailment score between the evidence and each candidate correction. The highest-scoring correction is selected as the final output. %
An overview of our framework is shown in \Cref{fig:framework_overview}. Our method ensures the corrected information units are derived from the evidence, which helps improve the faithfulness of the generated corrections. In addition, our approach is \textit{naturally interpretable} since the questions and answers generated directly reflect which information units are being compared with the evidence.\looseness=-1

Our contributions can be summarized as follows:
\begin{itemize}[noitemsep,nolistsep]
  \item We propose \modelshort~, a factual error correction framework that effectively addresses faithfulness by asking questions about the input claim, seeking answers in the evidence, and scoring the outputs by faithfulness.
  \item Our approach outperforms all prior methods, including fully-supervised approaches trained on 58K instances, in ensuring faithfulness on two factual error correction datasets, \textsc{FEVER} \cite{thorne-etal-2018-fever} and \textsc{Scifact} \cite{wadden-etal-2020-fact}. %
  \item We analyze the correlation of human judgments with automatic metrics to provide intuition for future research on evaluating the faithfulness, factuality, and intelligibility of factual error corrections.
  
\end{itemize}

\section{Task}

In \citet{thorne-vlachos-2021-evidence}'s setting, retrieved evidence is used, which means the model may be able to correct factual errors, even though there is no supporting information in the evidence. In this case, although the prediction is considered correct, the model is hallucinating, which is not a desired property. Additionally, due to the way data was collected, they require  systems to alter the input claim even if the input claim is already faithful to the evidence. We argue that \textit{no edit is needed for claims that are faithful to the evidence}.\looseness=-1 %

To address these shortcomings, our setup aims to edit a claim using a given piece of grounded evidence that supports or refutes the original claim (see \Cref{fig:framework_overview}). Using gold-standard evidence avoids the issue where a system outputs the correct answer by chance due to hallucinations. In our setting, a system must be faithful to the evidence to correct factual errors, allowing us to evaluate system performance more fairly. Furthermore, we require the model not to edit the original claim if it is already factually consistent with the provided evidence. \looseness=-1

Concretely, the input to our task is a claim $\mathcal{C}$ and a piece of gold-standard evidence $\mathcal{E}$ that supports or refutes $\mathcal{C}$. The goal of factual error correction is to produce a corrected claim $\hat{\mathcal{C}}$ that fixes factual errors in $\mathcal{C}$ while being faithful to $\mathcal{E}$. If $\mathcal{C}$ is already supported by $\mathcal{E}$, models should output the original claim (i.e. $\hat{\mathcal{C}} = \mathcal{C}$).
\section{Proposed Methods}
Our framework, \modelshort~, faithfully corrects factual errors using question-answering and entailment. Specifically, we represent the input claim $\mathcal{C}$ as question-answer pairs $\{ (Q_1, A^{\mathcal{C}}_1), ..., (Q_n, A^{\mathcal{C}}_n)\}$ such that each question $Q_i$ reflects the corresponding information unit $A^{\mathcal{C}}_i$, such as noun phrases and adjectives (\Cref{subsec:candidate_gen} and \Cref{subsec:question_gen}). Based on each question $Q_i$, we look for an answer $A^{\mathcal{E}}_i$ in the given evidence $\mathcal{E}$ using a learned QA model (\Cref{subsec:question_ans}). Each candidate correction $S_i$ is obtained by converting the corresponding pair of $Q_i$ and $A^{\mathcal{E}}_i$ into a declarative statement (\Cref{subsec:qa_to_claim}). This guarantees that the corrected information units we replace factual errors with are derived from the evidence and thus ensures high faithfulness. The final output of \modelshort~ is the $S_i$ with the highest faithfulness score computed by an entailment model (\Cref{subsec:correction_ranking}). An overview of our framework is shown in \Cref{fig:framework_overview}.\looseness=-1

One major challenge that makes our task more difficult than prior studies on faithfulness \cite{wang-etal-2020-asking, fabbri-etal-2022-qafacteval} is that we need to handle more diverse factual errors, such as negation errors and errors that can only be abstractively corrected. For instance, in the second example of in \Cref{tab:qualitative}, the QA model should output ``Yes'' as the answer, which cannot be produced by extractive QA systems. To address this issue, we adopt abstractive QG and QA models that can handle diverse question types and train our QA-to-claim model on multiple datasets to cover cases that cannot be handled by extractive systems. The following subsections illustrate the details of each component in our framework. %

\subsection{Claim Answer Generation}
\label{subsec:candidate_gen}
The goal of claim answer generation is to identify information units in the input claim that may be unfaithful to $\mathcal{E}$. We aim to maximize the recall in this step since the missed candidates cannot be recovered in later steps. Therefore, we extract all noun chunks and named entities using Spacy\footnote{\url{https://spacy.io/}} and extract nouns, verbs, adjectives, adverbs, noun phrases, verb phrases using Stanza\footnote{\url{https://stanfordnlp.github.io/stanza/}}. Additionally, we also extract negation terms, such as ``not'' and ``never'', %
from the input claim. We name the extracted information units \textit{claim answers}, denoted as $A^{\mathcal{C}} = \{ A^{\mathcal{C}}_1, A^{\mathcal{C}}_2, ..., A^{\mathcal{C}}_n\}$. %

\subsection{Question Generation}
\label{subsec:question_gen}
Upon \textit{claim answers} are produced, we generate questions that will be later used to look for correct information units in the evidence. Questions are generated conditioned on the \textit{claim answers} using the input claim as context. We denote the question generator as $\mathcal{G}$. Each \textit{claim answer} $A^{\mathcal{C}}_i$ is concatenated with the input claim $\mathcal{C}$ to generate a question $Q_i = \mathcal{G}(A^{\mathcal{C}}_i, \mathcal{C})$. We utilize MixQG \cite{murakhovska-etal-2022-mixqg} as our question generator $\mathcal{G}$ to cover the wide diversity of factual errors and candidates extracted. MixQG was trained on nine question generation datasets with various answer types, including boolean, multiple-choice, extractive, and abstractive answers. %

\subsection{Question Answering}
\label{subsec:question_ans}
The question answering step identifies the correct information units $A^{\mathcal{E}}_i$ corresponding to each question $Q_i$ in the given evidence $\mathcal{E}$. Our QA module answers questions from the question generation steps with the given evidence as context. Let $\mathcal{F}$ denote our QA model. We feed the concatenation of a generated question and the evidence to the QA model to produce an \textit{evidence answer} $A^{\mathcal{E}}_i = \mathcal{F}(Q_i, \mathcal{E})$. UnifiedQA-v2 \cite{khashabi2022unifiedqa} is used as our question answering model. UnifiedQA-v2 is a T5-based \cite{raffel2020exploring} abstractive QA model trained on twenty QA datasets that can handle diverse question types. %

\subsection{QA-to-Claim}
\label{subsec:qa_to_claim}
After questions and answers are generated, we transform each pair of question and answer into a declarative statement, which serves as a candidate correction that will be scored in the next step. Previous studies on converting QAs to claims focus on extractive answer types only \cite{pan-etal-2021-zero}. To accommodate diverse types of questions and answers, we train a sequence-to-sequence model that generates a claim given a question-answer pair on three datasets: QA2D \cite{demszky2018transforming} for extractive answers, BoolQ \cite{clark-etal-2019-boolq} for boolean answers, and SciTail \cite{scitail} for covering scientific domain QAs. Note that samples in BoolQ do not contain converted declarative statements. Using Stanza's constituency parser, we apply heuristics to transform all QAs to their declarative forms in BoolQ. Our QA-to-claim model is a T5-base fine-tuned on these three datasets. Concretely, let $\mathcal{M}$ denote our QA-to-claim model. $\mathcal{M}$ takes in a \textit{generated question} $\mathcal{Q}_i$ and an \textit{evidence answer} $A^{\mathcal{E}}_i$ as inputs and outputs a statement $S_i = \mathcal{M}(Q_i, A^{\mathcal{E}}_i)$.

\subsection{Correction Scoring}
\label{subsec:correction_ranking}
The final correction is produced by scoring the faithfulness of each candidate correction from the previous steps w.r.t. the evidence. We use entailment score to approximate faithfulness. Here, DocNLI \cite{yin-etal-2021-docnli} is used to compute such document-sentence entailment relations.
DocNLI is more generalizable than other document-sentence entailment models, such as FactCC \cite{kryscinski-etal-2020-evaluating}, since it was trained on five datasets of various tasks and domains. Conventional NLI models trained on sentence-level NLI datasets, such as MNLI \cite{williams-etal-2018-broad}, are not applicable since previous work has found that these models are ill-suited for measuring entailment beyond the sentence level \cite{falke-etal-2019-ranking}. In addition, to prevent the final correction from deviating too much from the original claim, we also consider ROUGE-1 scores, motivated by \citet{wan-bansal-2022-factpegasus}. The final metric used for scoring is the sum of ROUGE-1 score\footnote{\url{https://pypi.org/project/py-rouge/}} and DocNLI entailment score. Formally, 

{
\small~
\begin{align}
     \mathcal{V}(S_i) &= \text{DocNLI}(S_i, \mathcal{E}) + \text{ROUGE-1}(S_i, \mathcal{C}) \\
     \mathcal{C}' &= \argmax_{S_i} \mathcal{V}(S_i), 
\end{align}
}
where $\mathcal{C}'$ is the final correction produced by our framework.
Furthermore, to handle cases where the input claim is already faithful to the evidence, we include the input claim in the candidate correction list to be scored.

\subsection{Domain Adaptation}
During the early stage of our experiments, we found that our proposed framework did not perform well in correcting factual errors in biomedical claims. This results from the fact that our QA and entailment models were not fine-tuned on datasets in the biomedical domain. To address this issue, we adapt \textsc{UnifiedQA-v2} and \textsc{DocNLI} on two biomedical QA datasets, \textsc{PubMedQA} \cite{jin-etal-2019-pubmedqa} and \textsc{BioASQ} \cite{tsatsaronis2015overview}, by further fine-tuning them for a few thousand steps. We later show that this simple domain adaptation technique successfully improves our overall factual error correction performance on a biomedical dataset without decreasing performance in the Wikipedia domain (see \Cref{subsec:main_results}). %

\section{Experimental Setup}

\subsection{Datasets}
We conduct experiments on two English datasets, \textsc{FEVER} and \textsc{SciFact}. \textsc{FEVER} \cite{thorne-vlachos-2021-evidence} is repurposed from the corresponding fact-checking dataset \cite{thorne-etal-2018-fever} that consists of evidence collected from Wikipedia and claims written by humans that are supported or refuted by the evidence. %
Similarly, \textsc{SciFact} is another fact-checking dataset in the biomedical domain \cite{wadden-etal-2020-fact}. We repurpose it for the factual error correction task using the following steps. First, we form faithful claims by taking all claims supported by evidence. Then, unfaithful claims are generated by applying Knowledge Base Informed Negations \cite{wright-etal-2022-generating}, a semantic altering transformation technique guided by knowledge base, to a subset of the faithful claims. \Cref{apx:datset_stats} shows detailed statistics. \looseness=-1

\subsection{Evaluation Metrics}
Our evaluation focuses on faithfulness. Therefore, we adopt some recently developed metrics that have been shown to correlate well with human judgments in terms of faithfulness. BARTScore \cite{yuan2021bartscore} computes the semantic overlap between the input claim and the evidence by calculating the logarithmic probability of generating the evidence conditioned on the claim. FactCC \cite{kryscinski-etal-2020-evaluating} is an entailment-based metric that predicts the faithfulness probability of a claim w.r.t. the evidence. We report the average of the \textsc{Correct} probability across all samples. In addition, we consider \textsc{QAFactEval} \cite{fabbri-etal-2022-qafacteval}, a recently released QA-based metric that achieves the highest performance on the \textsc{SummaC} factual consistency evaluation benchmark \cite{laban-etal-2022-summac}. Furthermore, we also report performance on SARI \cite{Xu-EtAl:2016:TACL}, a lexical-based metric that has been widely used in the factual error correction task \cite{thorne-vlachos-2021-evidence, shah2020automatic}. \looseness=-1 %

\subsection{Baselines}
We compare our framework with the following baseline systems. \textbf{\textsc{T5-Full}} \cite{thorne-vlachos-2021-evidence} is a fully-supervised model based on T5-base \cite{2020t5} that generates the correction conditioned on the input claim and the given evidence. \textbf{\textsc{MaskCorrect}} \cite{shah2020automatic} and \textbf{\textsc{T5-Distant}} \cite{thorne-vlachos-2021-evidence} are both distantly-supervised methods that are composed of a masker and a sequence-to-sequence (seq2seq) corrector. The masker learns to mask out information units that are possibly false based on a learned fact verifier or an explanation model \cite{ribeiro-etal-2016-why} and the seq2seq corrector learns to fill in the masks with factual information. The biggest difference is in the choice of seq2seq corrector. \textbf{\textsc{T5-Distant}} uses T5-base, while \textbf{\textsc{MaskCorrect}} utilizes a two-encoder pointer generator. For zero-shot baselines, we selected two post-hoc editing frameworks that are trained to remove hallucinations from summaries, \textbf{\textsc{ReviseRef}} \cite{adams2022learning} and \textbf{\textsc{CompEdit}} \cite{fabbri2022improving}. \textbf{\textsc{ReviseRef}} is trained on synthetic data where hallucinating samples are created by entity swaps. \textbf{\textsc{CompEdit}} learns to remove factual errors with sentence compression, where training data are generated with a separate perturber that inserts entities into faithful sentences. %

\subsection{Implementation Details}
No training is needed for \modelshort~. As for \modelshortda~, we fine-tune \textsc{UnifiedQA-v2} and \textsc{DocNLI} on the \textsc{BioASQ} and \textsc{PubMedQA} datasets for a maximum of 5,000 steps using AdamW \cite{loshchilov2018decoupled} with a learning rate of 3e-6 and a weight decay of 1e-6. During inference time, all generative components use beam search with a beam width of 4.

\section{Results}

\subsection{Main Results}
\label{subsec:main_results}

\begin{table*}[t]
    \small
    \centering

    \begin{tabularx}{\textwidth}{l *{8}{Y}}
        \toprule
        
        \multirow{2}{*}{\textbf{Method}}  & \multicolumn{4}{c}{\textbf{\textsc{FEVER}}} & \multicolumn{4}{c}{\textbf{\textsc{SciFact}}} \\
        \cmidrule(lr){2-5}
        \cmidrule(lr){6-9}
        
           &  \textsc{SARI}~(\%)   &  \textsc{BS}  &  \textsc{QFE}  &  \textsc{FC}~(\%)    &  \textsc{SARI}~(\%)    &  \textsc{BS}  &  \textsc{QFE}  &  \textsc{FC}~(\%)  \\ 
        \midrule
        \multicolumn{9}{c}{\textit{Fully-supervised}} \\[2ex]

         \textsc{T5-Full}    & 35.50 & -2.74 & 1.40 & 41.91	& 35.07 & -3.12 & 1.23 & 50.17\\
        
        \midrule
        \multicolumn{9}{c}{\textit{Distantly-supervised}} \\[2ex]

         \textsc{MaskCorrect}  & 25.66 & -4.48 & 0.67 & 20.12  & 15.21 & -4.31 & 0.54 & 34.92 \\
        \textsc{T5-Distant}   & \hlc{lightgreen}{36.01}  & -2.90 & 1.12  & 32.28 & 20.08 & -3.51 & 0.99 &  44.77\\
        \midrule
        \multicolumn{9}{c}{\textit{Zero-shot}} \\[2ex]
        
        \textsc{ReviseRef}   & 20.52 & -5.27 & 0.30 & 26.00 & 17.53 & -4.58 & 0.97 & \hlc{lightgreen}{\textbf{52.44}}\\
        \textsc{CompEdit}   & 25.51 & -2.83 & 1.23 & 39.46 & 25.41 & -3.31 & 1.12 & \hlc{lightgreen}{50.62}\\
        \modelshort~ (Ours) & ~\hlc{lightgreen}{39.16}$^{*}$ & ~\hlc{lightgreen}{\textbf{-2.58}}$^{*}$ &  ~\hlc{lightgreen}{\textbf{2.06}}$^{*}$ & ~\hlc{lightgreen}{\textbf{47.08}}$^{*}$ & 29.67 & -3.22 & 1.12 & 47.84\\
        \modelshort~\textsc{-DA} (Ours) & ~\hlc{lightgreen}{\textbf{40.65}}$^{*}$ & ~\hlc{lightgreen}{-2.67}$^{*}$ & ~\hlc{lightgreen}{2.03}$^{*}$ & ~\hlc{lightgreen}{45.75}$^{*}$  & \textbf{31.93} &	\textbf{-3.21}  & ~\hlc{lightgreen}{\textbf{1.30}}$^{*}$ & 50.10 \\
        \bottomrule
    \end{tabularx}

    \caption{Main results on the \textsc{FEVER} and \textsc{SciFact} datasets. \textsc{BS} denotes \textsc{BARTScore}, \textsc{QFE} denotes \textsc{QAFactEval}, and \textsc{FC} denotes \textsc{FactCC}. \modelshortda~ is our framework with the QA and entailment components further fine-tuned on biomedical QA datasets. Among distantly-supervised and zero-shot results, the best scores per metric are marked in \textbf{boldface}. Models achieving performance better than the fully-supervised model are marked in \hlc{lightgreen}{gray}. Statistical significance over previous best methods computed with the paired bootstrap procedure \cite{berg-kirkpatrick-etal-2012-empirical} are indicated with $^*$ ($p< .01$). } 
    \label{tab:main}
\end{table*}
\Cref{tab:main} summarizes the main results on the \textsc{FEVER} and \textsc{SciFact} datasets. Both \modelshort~ and \modelshortda~ achieve significantly better performance than the distantly-supervised and zero-shot baselines. More impressively, they surpass the performance of the fully-supervised model on most metrics, even though the fully-supervised model is trained on 58K samples in the \textsc{FEVER} experiment. The improvements demonstrate the effectiveness of our approach in producing faithful factual error correction by combining question answering and entailment predictions. In addition, even though our domain adaptation technique is simple, it successfully boosts the performance on the \textsc{SciFact} dataset while pertaining great performance on the \textsc{FEVER} dataset. The first example in \Cref{tab:qualitative} illustrates an instance where domain adaptation fixes an error made by \modelshort~. The absence of domain adaptation results in incorrect predictions by \modelshort~, as DocNLI assigns a significantly lower entailment score to the correct candidate ``Clathrin stabilizes the spindle fiber apparatus during mitosis phase.'' and a higher score to the wrong candidate ``Clathrin stabilizes the spindle apparatus during anaphase?'', indicating poor entailment assessment. With domain adaptation, \modelshortda~ resolves this issue by enabling DocNLI to approximate faithfulness more accurately.

It is true that \modelshortda~ requires additional training, which is different from typical zero-shot methods. However, the key point remains that our framework does not require any task-specific training data. Hence, our approach still offers the benefits of zero-shot learning by not requiring any additional training data beyond what was already available for the question answering task, a field with much richer resources compared to the fact-checking field.

\subsection{Qualitative Analysis}
\begin{figure}[h]
    \centering
    \includegraphics[width=0.9\linewidth]{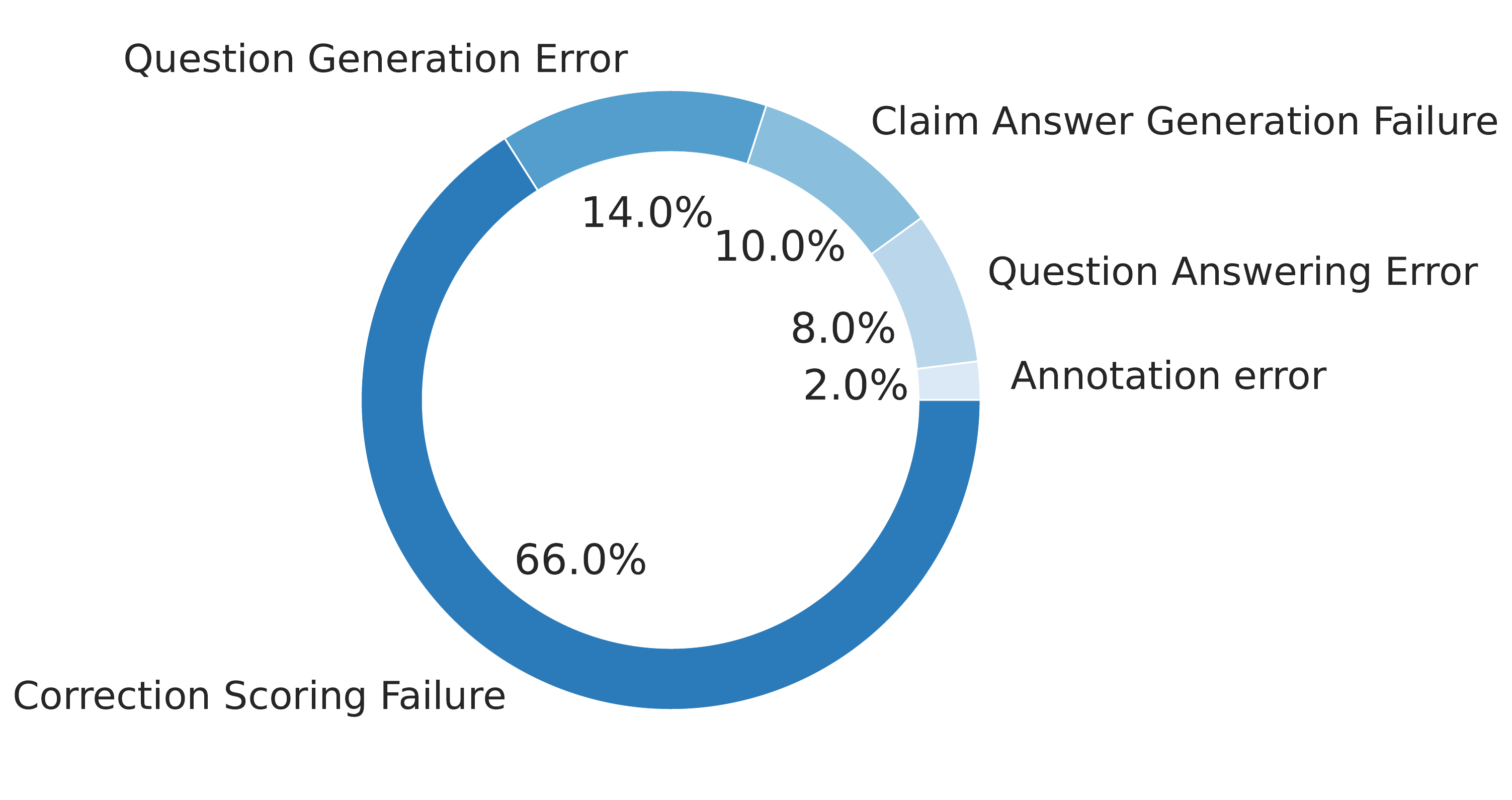}
    \vspace{-2mm}
    \caption{Distributions of errors.}  %
    \vspace{-5mm}
    
    \label{fig:error_analysis}
\end{figure}

To provide intuition for our framework’s ability to produce faithful factual error corrections, we manually examined 50 correct and 50 incorrect outputs made by \modelshort~ on the \textsc{FEVER} dataset. The interpretability of \modelshort~ allows for insightful examinations of the outputs. Among the correct samples, our framework produces faithful corrections because all intermediate outputs are accurately produced rather than ``being correct by chance''. For the incorrect outputs, we analyze the source of mistakes, as shown in \Cref{fig:error_analysis}. The vast majority of failed cases result from DocNLI's failure to score candidate corrections faithfully. In addition to the mediocre performance of DocNLI, one primary reason is that erroneous outputs from other components would not be considered mistakes so long as the correction scoring module determines the resulting candidate corrections unfaithful to the evidence. A possible solution to improve DocNLI is to further fine-tune it on synthetic data generated by perturbing samples in \textsc{FEVER} and \textsc{SciFact}. Examples of correct and incorrect outputs are presented in \Cref{tab:pos_examples} and \Cref{tab:neg_examples} of \Cref{apx:additional_qualitative}, respectively. \looseness=-1

\begin{table*}[t]
    \small
    \centering
    \begin{adjustbox}{max width=\textwidth}
    {
    \begin{tabular}{ll}
        
        \toprule
        \multicolumn{2}{c}{\textbf{Example 1}} \\

        \arrayrulecolor{black!30}\midrule
        \multicolumn{2}{l}{\textbf{Input claim}: Clathrin stabilizes the spindle fiber apparatus during anaphase.} \\
        \multicolumn{2}{l}{\textbf{Evidence}: ...but is shut down during mitosis, when clathrin concentrates at the spindle apparatus...}\\
        
        \multicolumn{2}{l}{\textbf{Gold correction}: Clathrin stabilizes the spindle fiber apparatus during mitosis.} \\
        \arrayrulecolor{black!30}\midrule
        \textbf{Claim answer}: anaphase & \textbf{Generated question}: Clathrin stabilizes the spindle fiber apparatus during what phase? \\
        \textbf{Evidence answer}: mitosis & \textbf{Candidate correction}: Clathrin stabilizes the spindle fiber apparatus during mitosis phase.\\
        \textbf{DocNLI + ROUGE-1}: 0.0165 + 0.8235 & \textbf{\modelshort~'s output}: \hlc{lightred}{Clathrin stabilizes the spindle apparatus during anaphase?}\\

        \arrayrulecolor{black!30}\midrule
        \textbf{Claim answer}: anaphase & \textbf{Generated question}: Clathrin stabilizes the spindle fiber apparatus during what phase? \\
        \textbf{Evidence answer}: mitosis & \textbf{Candidate correction}: Clathrin stabilizes the spindle fiber apparatus during mitosis phase. \\
        \textbf{DocNLI + ROUGE-1}: 0.9999 + 0.8235 & \textbf{\modelshortda~'s output}: \hlc{lightblue}{Clathrin stabilizes the spindle fiber apparatus during mitosis phase.}\\

        \arrayrulecolor{black}
        \midrule
        
        \multicolumn{2}{c}{\textbf{Example 2}} \\

        \arrayrulecolor{black!30}\midrule
        \multicolumn{2}{l}{\textbf{Input claim}: Fuller House (TV series) won't air on Netflix.} \\
        \multicolumn{2}{l}{\textbf{Evidence}: Fuller House is an American family sitcom and sequel to the 1987-95 television series Fuller House, airing as a Netflix original series...} \\
        \multicolumn{2}{l}{\textbf{Gold correction}: Fuller House (TV series) airs on Netflix.} \\
        \arrayrulecolor{black!30}\midrule
        \textbf{Claim answer}: won't air on Netflix & \textbf{Generated question}: Does Fuller House air on Netflix? \\
        \textbf{Evidence answer}: Yes & \textbf{Candidate correction}: Fuller House airs on Netflix. \\
        \textbf{DocNLI + ROUGE-1}: 0.7222 + 0.7143 & \textbf{\modelshort~'s output}: \hlc{lightblue}{Fuller House airs on Netflix.} \\
        
        \arrayrulecolor{black!30}\midrule
        
        \multicolumn{2}{l}{\textbf{\textsc{T5-distant}'s output}: \hlc{lightred}{Fuller House ( TV series ) isn't airing on HBO.}} \\
        
        \arrayrulecolor{black}\bottomrule
    \end{tabular}
    }
    \end{adjustbox}
    \caption{Example outputs from different approaches. The outputs from our framework are directly interpretable, as the generated questions and answers reflect which information units in the input claim are erroneous and which information in the evidence supports the final correction. We only show the generated answers and questions directly related to the gold correction. In the first example, \modelshortda~ corrects a mistake made by \modelshort~ thanks to domain adaptation.  In the second example, \modelshort~ successfully produces a faithful factual error correction, whereas the output of \textsc{T5-distant}, the distantly-supervised baseline, is factual yet unfaithful to the evidence. }  %
    
    \label{tab:qualitative}
\end{table*}

\subsection{Human Evaluation}
\label{subsec:human_eval}
To further validate the effectiveness of our proposed method, we recruited three graduate students who are not authors to conduct human evaluations on 100 and 40 claims from \textsc{FEVER} and \textsc{SciFact}, respectively. For each claim, human judges are presented with the ground-truth correction, the gold-standard evidence, and output produced by a factual error correction system and tasked to assess the quality of the correction with respect to three dimensions. \textit{\textbf{Intelligibility}} evaluates the fluency of the correction. An intelligible output is free of grammatical mistakes, and its meaning must be understandable by humans without further explanation. \textit{\textbf{Factuality}} considers whether the input claim is aligned with facts. Systems' output can be factual and semantically different from the gold correction as long as it is consistent with the world's knowledge. \textit{\textbf{Faithfulness}} examines whether the input is factually consistent with the given evidence. Note that a faithful output must be factual since we assume all evidence is free of factual error. To evaluate the annotation quality, we compute the inter-annotator agreement. Krippendorff's Alpha \cite{krippendorff2011computing} is 68.85\%, which indicates a moderate level of agreement. Details of our human evaluation can be found in \Cref{apx:human_eval}.

\begin{table}[b]
    \small
    \centering
    \begin{adjustbox}{max width=0.47\textwidth}
    {
    \begin{tabular}{lcccccc}
        \toprule
        
        \multirow{2}{*}{\textbf{Method}}  & \multicolumn{3}{c}{\textbf{\textsc{FEVER}}} & \multicolumn{3}{c}{\textbf{\textsc{SciFact}}} \\
        \cmidrule(lr){2-4}
        \cmidrule(lr){5-7}
        
           &  \textit{Intel.}   &  \textit{Fact.}  &  \textit{Faith.}    &  \textit{Intel.}   &  \textit{Fact.}  &  \textit{Faith.}    \\ 
        \midrule
        
         \textsc{T5-Full} & 0.983 & 0.516 & 0.509 & 0.972 & 0.683 & 0.610 \\
         \midrule 
         \textsc{T5-Distant}  & 0.891 & 0.471 & 0.412 & 0.628 & 0.186 & 0.116 \\
        \modelshort~  & \textbf{0.951} & \hlc{lightgreen}{0.797} & \hlc{lightgreen}{0.797} & 0.826 & 0.413 & 0.413 \\
        \modelshort~\textsc{-DA}  & 0.893 & \hlc{lightgreen}{\textbf{0.835}} & \hlc{lightgreen}{\textbf{0.835}} & \textbf{0.953} & \textbf{0.628} & \hlc{lightgreen}{\textbf{0.628}}~  \\
        \bottomrule
    \end{tabular}
    }
    \end{adjustbox}
    \caption{Human evaluation on the \textsc{FEVER} and \textsc{SciFact} datasets. \textit{Intel.} denotes \textit{Intelligibility},  \textit{Fact.} denotes \textit{Factuality}, and  \textit{Faith.} denotes \textit{Faithfulness}.} 
    \label{tab:human_eval}
\end{table}

The human evaluation results are demonstrated in \Cref{tab:human_eval}. We observe that: (1) \modelshort~ and \modelshortda~ achieve the best overall performance in \textit{Factuality} and \textit{Faithfulness} on both datasets, even when compared to the fully-supervised method, suggesting that our approach is the best in ensuring faithfulness for factual error correction. (2) Our domain adaptation for the biomedical domain surprisingly improves faithfulness and factuality in the Wikipedia domain (i.e. \textsc{FEVER}). This suggests that fine-tuning the components of our framework on more datasets helps improve robustness in terms of faithfulness. (3) Factual output produced by \modelshort~ and \modelshortda~ are always faithful to the evidence, preventing the potential spread of misleading information caused by factual but unfaithful corrections. The second example in \Cref{tab:qualitative} demonstrates an instance of factual but unfaithful correction made by baseline models. Here, the output of \textsc{T5-distant} is unfaithful since the evidence does not mention whether Fuller House airs on HBO. In fact, although Fuller House was not on HBO when it premiered, it was later accessible on HBO Max. Therefore, the correction produced by \textsc{T5-distant} is misleading.\looseness=-1 %

\subsection{Correlation with Human Judgments}
Recent efforts on faithfulness metrics have been mostly focusing on the summarization task. No prior work has studied the transferability of these metrics to the factual error correction task. We seek to bridge this gap by showing the correlation between the automatic metrics used in \Cref{tab:main} and the human evaluation results discussed in \Cref{subsec:human_eval}. Using Kendall's Tau \cite{kendall1938new} as the correlation measure, the results are summarized in \Cref{tab:correlation}. 

We have the following observations. (1) \textsc{SARI} is the most consistent and reliable metric for evaluating \textit{Factuality} and \textit{Faithfulness} across two datasets. Although the other three metrics developed more recently demonstrate high correlations with human judgments of faithfulness in multiple summarization datasets, their transferability to the factual error correction task is limited due to their incompatible design for this particular task. %
For example, QA-based metrics like \textsc{QAFactEval} are less reliable for evaluating faithfulness in this task due to their inability to extract a sufficient number of answers from a single-sentence input claim. In contrast, summaries in summarization datasets generally consist of multiple sentences, enabling the extraction of a greater number of answers. To validate this, we analyzed the intermediate outputs of \textsc{QAFactEval}. Our analysis confirms that it extracts an average of only 1.95 answers on the \textsc{FEVER} dataset, significantly lower than the more than 10 answers typically extracted for summaries. %
(2) Across the two datasets, the correlations between all automatic metrics and \textit{Intelligibility} are low. The extremely high proportion of intelligible outputs may explain the low correlation. (3) The correlation for learning-based metrics, including \textsc{QAFactEval} and \textsc{FactCC}, drop significantly when applied to \textsc{SciFact}. This is likely caused by the lack of fine-tuning or pre-training with biomedical data. \looseness=-1

\begin{table}[t]
    \small
    \centering
    \begin{adjustbox}{max width=0.47\textwidth}
    {
    \begin{tabular}{lcccccc}
        \toprule
        
        \multirow{2}{*}{\textbf{Metric}}  & \multicolumn{3}{c}{\textbf{\textsc{FEVER}}} & \multicolumn{3}{c}{\textbf{\textsc{SciFact}}} \\
        \cmidrule(lr){2-4}
        \cmidrule(lr){5-7}
        
           &  \textit{Intel.}   &  \textit{Fact.}  &  \textit{Faith.}    &  \textit{Intel.}   &  \textit{Fact.}  &  \textit{Faith.}    \\ 
        \midrule
        
         \textsc{SARI} & 0.017 & \textbf{0.370} & \textbf{0.383} & -0.026 & \textbf{0.379} & \textbf{0.412} \\
         \textsc{BARTScore}  & \textbf{0.137} & 0.071 & 0.104 & \textbf{0.104} & 0.118 & 0.119 \\
        \textsc{QAFactEval} & -0.045 & 0.360 & 0.379 & 0.084 & 0.234 & 0.272 \\
        \textsc{FactCC} & 0.053 & 0.203 & 0.225 & -0.119 & -0.073 & -0.076 \\
        
        \bottomrule
    \end{tabular}
    }
    \end{adjustbox}
    \caption{Correlation between automatic metrics and human judgments on the \textsc{FEVER} and \textsc{SciFact} datasets computed using Kendall's Tau.} 
    \label{tab:correlation}
\end{table}

\section{Related Work}

\subsection{Factual Error Correction}

An increasing number of work began to explore factual error correction in recent years, following the rise of fact-checking \cite{thorne-etal-2018-fever, wadden-etal-2020-fact, gupta-srikumar-2021-x, huang-etal-2022-concrete} and fake news detection \cite{shu2020fakenewsnet, fung-etal-2021-infosurgeon, wu-etal-2022-cross, huang2022faking}. \citet{shah2020automatic} propose a distant supervision learning method based on a masker-corrector architecture, which assumes access to a learned fact verifier. \citet{thorne-vlachos-2021-evidence} created the first factual error correction dataset by repurposing the FEVER \cite{thorne-etal-2018-fever} dataset, which allows for fully-supervised training of factual error correctors. They also extended \citet{shah2020automatic}'s method with more advanced pre-trained sequence-to-sequence models.
Most recently, \citet{schick2022peer} proposed \textsc{PEER}, a collaborative language model that demonstrates superior text editing capabilities due to its multiple text-infilling pre-training objectives, such as planning and realizing edits as well as explaining the intention behind each edit\footnote{We are not able to compare with \textsc{PEER} \cite{schick2022peer} as its checkpoints have not been released by the time we ran the experiments.}. %

\subsection{Faithfulness}%
Previous studies addressing faithfulness are mostly in the summarization field and can be roughly divided into two categories, evaluation and enhancement. Within faithfulness evaluation, one line of work developed entailment-based metrics by training document-sentence entailment models on synthetic data \cite{kryscinski-etal-2020-evaluating, yin-etal-2021-docnli} or human-annotated data~\cite{DBLP:conf/naacl/factgraph22,finegrainfact}, or applying conventional NLI models at the sentence level \cite{laban-etal-2022-summac}. Another line of work evaluates faithfulness by comparing information units extracted from summaries and input sources using QA \cite{wang-etal-2020-asking, deutsch-etal-2021-towards}. There is a recent study that integrates QA into entailment by feeding QA outputs as features to an entailment model \cite{fabbri-etal-2022-qafacteval}. %
We combine QA and entailment by using entailment to score the correction candidates produced by QA. %

Within faithfulness enhancement, some work improves factual consistency by incorporating auxiliary losses into the training process \cite{nan-etal-2021-improving, cao-wang-2021-cliff, tang-etal-2022-confit, huang-etal-2023-swing}. Some other work devises factuality-aware pre-training and fine-tuning objectives to reduce hallucinations \cite{wan-bansal-2022-factpegasus}. The most similar to our work are studies that utilize a separate rewriting model to fix hallucinations in summaries. For example, \citet{cao-etal-2020-factual} present a post-hoc corrector trained on synthetic data, where negative samples are created via perturbations. \citet{adams2022learning} fix factually inconsistent information in the reference summaries to prevent the summarization from learning hallucinating examples. \citet{fabbri2022improving} propose a compression-based post-editor to correct extrinsic errors in the generated summaries. By contrast, we leverage the power of QA and entailment together to address faithfulness.

\section{Conclusions and Future Work}
We have presented \modelshort~, a zero-shot framework that asks questions about an input claim and seeks answers from the given evidence to correct factual errors faithfully. The experimental results demonstrate the superiority of our approach over prior methods, including fully-supervised methods, as indicated by both automatic metrics and human evaluations. More importantly, the decomposability of \modelshort~ naturally offers interpretability, as the questions and answers generated directly reflect which information units in the input claim are incorrect and why. Furthermore, we reveal the most suitable metric for assessing faithfulness of factual error correction by analyzing the correlation between the reported automatic metrics and human judgments. For future work, we plan to extend our framework to faithfully correct misinformation in social media posts and news articles to inhibit the dissemination of false information. In addition, it may be meaningful to explore extending zero-shot factual error correction to multimedia task settings, such as identifying inconsistencies between chart and text \cite{zhou2023chart}.
\section{Limitations}
Although our approach has demonstrated advantages in producing faithful factual error corrections, we recognize that our approach is not capable of correcting all errors, particularly those that require domain-specific knowledge, as illustrated in \Cref{tab:human_eval}. Therefore, it is important to exercise caution when applying this framework in user-facing settings. For instance, end users should be made aware that not all factual errors may be corrected.

In addition, our approach assumes evidence is given. Although this assumption is also true for applying our method to summarization tasks since the source document is treated as evidence, it does not hold for automatic textual knowledge base updates. When updating these knowledge bases, it is often required to retrieve relevant evidence from external sources. Hence, a reliable retrieval system is required when applying our method to this task.

\section{Ethical Considerations}

While no fine-tuning is needed for \modelshort~, its inference time and memory usage are three to four times more than similar-sized baseline systems due to its multi-component architecture, implying higher environmental costs during test time. In addition, the underlying components of our method are based on language models pre-trained on data collected from the internet. These language models have been shown to exhibit potential issues, such as political or gender biases. While we did not observe such biases during our experiments, users of these models should be aware of these issues when applying them.\looseness=-1
\section*{Acknowledgement}
This research is based upon work supported by U.S. DARPA SemaFor Program No. HR001120C0123, DARPA AIDA Program No. FA8750-18-2-0014, and DARPA MIPs Program No. HR00112290105. The views and conclusions contained herein are those of the authors and should not be interpreted as necessarily representing the official policies, either expressed or implied, of DARPA, or the U.S. Government. The U.S. Government is authorized to reproduce and distribute reprints for governmental purposes notwithstanding any copyright annotation therein. 
Hou Pong Chan was supported in part by the Science and Technology Development Fund, Macau SAR (Grant Nos. FDCT/060/2022/AFJ, FDCT/0070/2022/AMJ) and the Multi-year Research Grant from the University of Macau (Grant No. MYRG2020-00054-FST). 
\looseness=-1
\bibliography{anthology,custom}
\bibliographystyle{acl_natbib}

\clearpage
\appendix
\section{Dataset Statistics}
\label{apx:datset_stats}
Details of the dataset statistics are shown in \Cref{tab:dataset_stats}.

\begin{table}[h]
    \small
    \centering
    \begin{adjustbox}{max width=0.48\textwidth}
    {
    \begin{tabular}{cccc}
        \toprule
        
        Dataset & \# Test Samples & \# \textsc{Supports} & \# \textsc{Refutes} \\
        \midrule
        \textsc{FEVER} & 3,882 & 1,593 & 2,289  \\
        \textsc{SciFact} & 100 & 43 & 57 \\
        \bottomrule
    \end{tabular}
    }
    \end{adjustbox}
    
    \caption{Statistics of \textsc{FEVER} and \textsc{SciFact}. }
    \label{tab:dataset_stats}
    
\end{table}

\section{Human Evaluation Details}
\label{apx:human_eval}
\begin{figure*}[bt]
    \centering
    \includegraphics[width=0.9\linewidth]{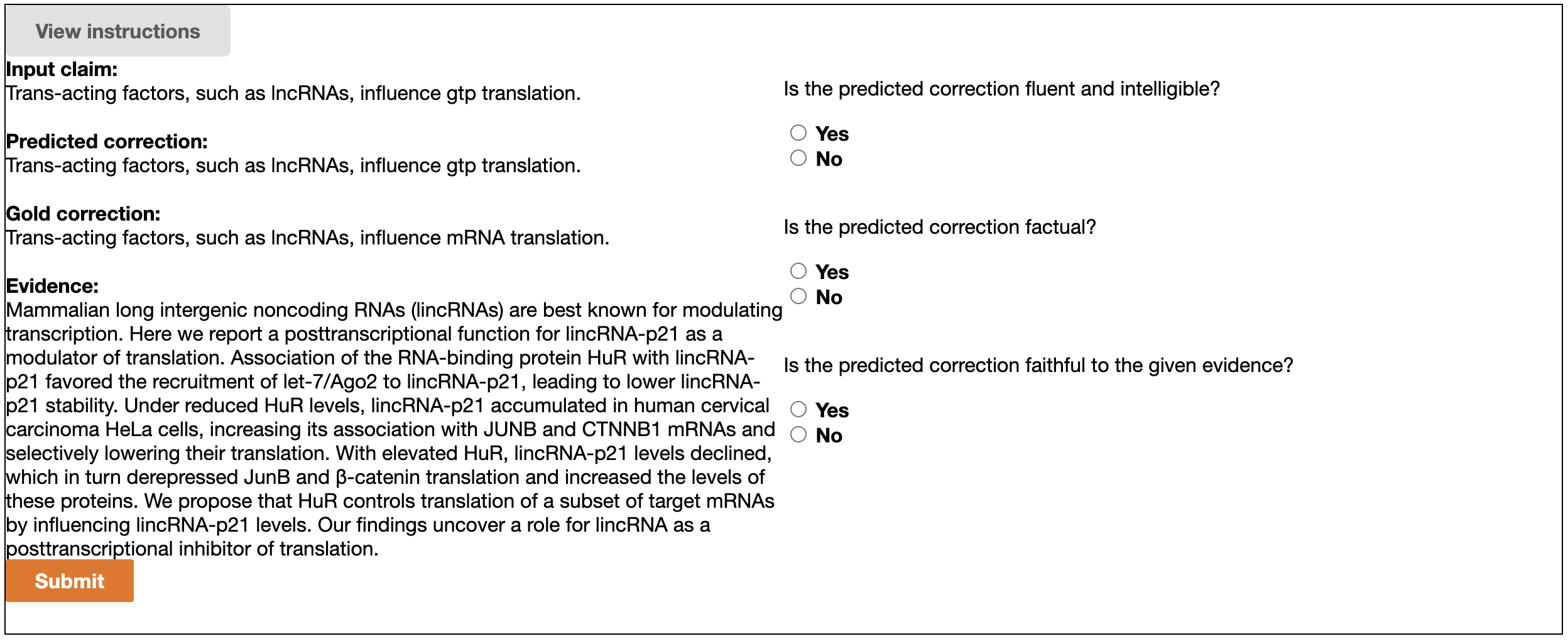}
    \vspace{-2mm}
    \caption{MTurk UI for our human evaluation.}
    \vspace{-5mm}
    \label{fig:mturk_ui}
\end{figure*}

In this section, we describe the details of our human evaluation. We recruit three engineering and science graduate students to ensure high-quality evaluation. For each HIT, annotators are provided with an input claim, the corresponding evidence and gold correction, and a predicted correction generated by a model. Based on the presented predictions, annotators are tasked to answer three questions shown on the right segment of the interface, each of which corresponds to \textit{Intelligence}, \textit{Factuality}, and \textit{Faithfulness}. They need to determine whether the predicted correction meets the three criteria according to each prompt. Our human evaluation interface is displayed in \Cref{fig:mturk_ui}.

Since the evaluation questions are self-explanatory, we only provide the human evaluators with terminology definitions and multiple examples of how evaluations should be conducted. Terminology is defined as follows:
\begin{itemize}
    \item \textbf{Input claim}: A sentence fed into a factual error correction system.
    \item \textbf{Predicted correction}: The output from the factual error correction system.
    \item \textbf{Gold correction}: Ground-truth label that the system aims to produce.
    \item \textbf{Evidence}: A document that the factual error correction system used to fix factual errors.
\end{itemize}

We maintain frequent communication with the human evaluators, including answering any questions they may have, to facilitate the evaluation process.

\begin{table}[t]
    \small
    \centering
    \begin{adjustbox}{max width=0.48\textwidth}
    {
    \begin{tabular}{lcc}
        \toprule

        \textbf{Model/Data Choice}   &  \textsc{SARI} (\%)   &  \textsc{QAFactEval}    \\
        \midrule
        \multicolumn{3}{c}{Question Generation} \\
        
        \midrule
         \textbf{MixQG-base} & 39.16 & \textbf{2.06}  \\
         T5-base (SQuAD) & \textbf{39.19} & 2.04 \\
        \midrule
        \multicolumn{3}{c}{Question Answering} \\
        \midrule

         \textbf{UnifiedQA-v2-base}  & \textbf{39.16} & 2.06  \\
         UnifiedQA-base  & 39.02 & \textbf{2.09} \\
         T5-base (SQuAD)  &  30.38 & 1.02 \\
         RoBERTa-base (SQuAD)  & 31.42 & 1.11 \\
        \midrule
        \multicolumn{3}{c}{QA-to-claim} \\
        \midrule
        
        \textbf{T5-base (QA2D + BoolQ + SciTail)}  & \textbf{39.16} & \textbf{2.06}  \\
        T5-base (QA2D) &  30.54 & 1.23  \\
        T5-base (SciTail) &  29.24 & 1.19  \\
        \midrule 
        \multicolumn{3}{c}{Correction Scoring} \\
        \midrule
        \textbf{DocNLI + ROUGE-1} & \textbf{39.16} & \textbf{2.06} \\
        DocNLI & 34.56 & 1.95\\
        FactCC + ROUGE-1 & 30.54 & 1.47 \\
        FactCC & 30.33 & 1.45 \\
        
        \bottomrule
    \end{tabular}
    }
    \end{adjustbox}
    \caption{Ablation studies on the \textsc{FEVER} dataset. The model used in \modelshort~ is \textbf{bolded}.} 
    \label{tab:ablation}
\end{table}
\section{Ablation Studies}
To understand how each component contributes to the performance of \modelshort~, we conducted ablation studies by replacing a given component in \modelshort~ with other models while keeping all other components the same as \modelshort~. We report the performance on the \textsc{FEVER} dataset in \textsc{SARI} and \textsc{QAFactEval} since these two metrics demonstrate the highest correlation with human judgments regarding faithfulness. Ablation results are presented in \Cref{tab:ablation}.

\paragraph{Effect of Question Generation}
We compared MixQG with a T5-base model trained on SQuAD \cite{rajpurkar-etal-2016-squad}. The results indicate that the final performance is not significantly affected by the use of either model. Upon further investigation, we surprisingly discovered that despite SQuAD exclusively comprising extractive question answering examples, the T5-base trained on it could generalize to other answer types. For example, given an answer ``not'' and a claim ``Cleopatre is not a queen.'', T5-base (SQuAD) generates ``Is Cleopatre a queen?''. Therefore, the training of MixQG on multiple QA datasets does not yield advantages.

\paragraph{Effect of Question Answering}
We experimented with an abstractive QA model, UnifiedQA \cite{khashabi-etal-2020-unifiedqa}, and two extractive QA models trained on SQuAD. We found that UnifiedQA performs similarly to UnifiedQA-v2, whereas using both extractive QA models leads to significant performance drops. This is likely due to the fact that SQuAD only includes extractive answer types. Although the encoder-decoder architecture of T5-base allows it to output words that do not present in the context, it fails to generate these types of answers. For instance, given a question ``Was Cleopatre a queen.'' and a context ``Cleopatra VII Philopator was Queen of the Ptolemaic Kingdom of Egypt...'', T5-base (SQuAD) would output ``Queen'' instead of ``Yes''.

\paragraph{Effect of QA-to-claim}
For QA-to-claim, we ablated different training data while keeping the same model architecture. Similar to our findings in the ablation studies on QA, when T5-base is only trained on QA2D or SciTail, it cannot convert boolean-typed questions and answers to declarative sentences, resulting in a marked decline in performance.

\paragraph{Effect of Correction Scoring}
We studied other scoring methods, including replacing DocNLI with FactCC and removing ROUGE-1. Using FactCC leads to a great performance drop, suggesting that DocNLI is likely a better approximation of faithfulness than FactCC. Furthermore, incorporating ROUGE-1 into the scoring criteria allows us to select a faithful correction that is most relevant to the input claim. Thus, we observe a huge drop in \textsc{SARI} when ROUGE-1 is removed.

\section{Additional Qualitative Analysis}
\label{apx:additional_qualitative}
As mentioned in \Cref{subsec:main_results}, we analyzed 50 correct and 50 incorrect outputs produced by \modelshort~. All 50 correct outputs are generated by asking the correct questions, answering correctly using the evidence, and scoring faithfully w.r.t. the evidence. Examples are demonstrated in \Cref{tab:pos_examples}.  For incorrect outputs, most of the errors are caused by DocNLI's inability to approximate faithfulness, as shown by the last instance in \Cref{tab:neg_examples}, even though DocNLI is the state-of-the-art document-sentence entailment model. In addition, annotation errors occur due to how the \textsc{FEVER} dataset was constructed (i.e. for fact-checking purposes). As demonstrated by the first example in \Cref{tab:neg_examples}, our correction is faithful to the evidence, and it is also more relevant to the input claim compared to the ground truth. As for errors in the question answering module, most of them are under-specified answers. For example, in the second instance in \Cref{tab:neg_examples}, the generated answer ``pop music duo'' is faithful to the evidence but is under-specified compared to the expected answer ``R\&B singers''.

\begin{table*}[t]
    \small
    \centering
    \begin{adjustbox}{max width=\textwidth}
    {
    \begin{tabular}{ll}
        
        \toprule

        \multicolumn{2}{l}{\textbf{Input claim}: University of Chicago Law School is ranked first in the 2016 QS World University Rankings.} \\
        \multicolumn{2}{l}{\textbf{Evidence}: The University of Chicago Law School is the graduate school of law at the University of Chicago. It is ranked 12th in the 2016 QA World University Rankings.}\\
        
        \multicolumn{2}{l}{\textbf{Gold correction}: University of Chicago Law School is ranked 12th in the 2016 QS World University Rankings.} \\
        \arrayrulecolor{black!20}\midrule
        \textbf{Claim answer}: ranked & \textbf{Generated question}: How is the University of Chicago Law School ranked in the 2016 QS World University Rankings? \\
        \textbf{Evidence answer}: 12th & \textbf{Candidate correction}: The University of Chicago Law School is 12th ranked in the 2016 QS World University Rankings.\\
        \textbf{DocNLI + ROUGE-1}: 0.8867 + 0.9032 & \textbf{\modelshort~'s output}: \hlc{lightblue}{The University of Chicago Law School is 12th ranked in the 2016 QS World University Rankings.}\\

        \arrayrulecolor{black} \midrule

        \multicolumn{2}{l}{\textbf{Input claim}: Simon Pegg was born on February 14th, 1860.} \\
        \multicolumn{2}{l}{\textbf{Evidence}: Fuller House is an American family sitcom and sequel to the 1987-95 television series Full House, airing as a Netflix original series ...} \\
        \multicolumn{2}{l}{\textbf{Gold correction}: Simon Pegg was born on February 14th, 1970.} \\
        \arrayrulecolor{black!20}\midrule
        \textbf{Claim answer}: February 14th, 1860 & \textbf{Generated question}: When was Simon Pegg born? \\
        \textbf{Evidence answer}: 14 february 1970 & \textbf{Candidate correction}: Simon Pegg was born on 14 february 1970. \\
        \textbf{DocNLI + ROUGE-1}: 0.9636 + 0.7500 & \textbf{\modelshort~'s output}: \hlc{lightblue}{Simon Pegg was born on 14 february 1970.} \\
        
        \arrayrulecolor{black} \midrule

        \multicolumn{2}{l}{\textbf{Input claim}: Caesar is a 1937 adaptation of The Tempest.} \\
        \multicolumn{2}{l}{\textbf{Evidence}: Fuller Caesar is the title of Orson Welles's innovative 1937 adaptation of William Shakespeare's Julius Caesar ...} \\
        \multicolumn{2}{l}{\textbf{Gold correction}: Caesar is a 1937 adaptation of Julius Caesar.} \\
        \arrayrulecolor{black!20}\midrule
        \textbf{Claim answer}: Tempest & \textbf{Generated question}: Caesar is a 1937 adaptation of what? \\
        \textbf{Evidence answer}: William Shakespeare's Julius Caesar & \textbf{Candidate correction}: Caesar is a 1937 adaptation of William Shakespeare's Julius Caesar. \\
        \textbf{DocNLI + ROUGE-1}: 0.9649 + 0.6315 & \textbf{\modelshort~'s output}: \hlc{lightblue}{Caesar is a 1937 adaptation of William Shakespeare's Julius Caesar.} \\

        \arrayrulecolor{black}\bottomrule
    \end{tabular}
    }
    \end{adjustbox}
    \vspace{-2mm}
    \caption{Example outputs from \modelshort~ on the \textsc{FEVER} dataset where it makes faithful factual error corrections.}  
    
    \label{tab:pos_examples}
\end{table*}
\begin{table*}[t]
    \small
    \centering
    \begin{adjustbox}{max width=\textwidth}
    {
    \begin{tabular}{ll}
        
        \toprule

        \multicolumn{2}{c}{\textbf{Annotation Error}} \\

        \arrayrulecolor{black!30}\midrule
        \multicolumn{2}{l}{\textbf{Input claim}: Ragtime was released in 1965.} \\
        \multicolumn{2}{l}{\textbf{Evidence}: Ragtime is a 1981 American drama film, ... The film features the final film appearances of James Cagney and Pat O'Brien ...}\\
        
        \multicolumn{2}{l}{\textbf{Gold correction}: Ragtime is the final film appearance of James Cagney.} \\
        \arrayrulecolor{black!20}\midrule
        \textbf{Claim answer}: released in 1965 & \textbf{Generated question}: \markred{When was Ragtime released?}~ \\
        \textbf{Evidence answer}: 1981 & \textbf{Candidate correction}: Ragtime was released in 1981. \\
        \textbf{DocNLI + ROUGE-1}: 0.9636 + 0.7500 & \textbf{\modelshort~'s output}: \hlc{lightred}{Ragtime was released in 1981.} \\
        \arrayrulecolor{black} \midrule
        
        \multicolumn{2}{c}{\textbf{Under-specified Answer}} \\

        \arrayrulecolor{black!30}\midrule
        \multicolumn{2}{l}{\textbf{Input claim}: Sonny \& Cher are rock singers.} \\
        \multicolumn{2}{l}{\textbf{Evidence}: Sonny \& Cher were an American pop music duo ... started their career in the mid-1960s as R\&B backing singers ...} \\
        \multicolumn{2}{l}{\textbf{Gold correction}: Sonny \& Cher are R\&B singers.} \\
        \arrayrulecolor{black!20}\midrule
        \textbf{Claim answer}: rock singers & \textbf{Generated question}: Sonny \& Cher are what type of singers? \\
        \textbf{Evidence answer}: \markred{pop music duo}~ & \textbf{Candidate correction}: Sonny \& Cher are a pop music duo. \\
        \textbf{DocNLI + ROUGE-1}: 0.8166 + 0.5000 & \textbf{\modelshort~'s output}: \hlc{lightred}{Sonny \& Cher are a pop music duo.} \\
        
        \arrayrulecolor{black} \midrule
        \multicolumn{2}{c}{\textbf{Correction Scoring Failure}} \\

        \arrayrulecolor{black!30}\midrule
        
        \multicolumn{2}{l}{\textbf{Input claim}: Johann Wolfgang von Goethe failed to publish Wilhelm meister's Apprenticeship.} \\
        \multicolumn{2}{l}{\textbf{Evidence}: ... During this period,  Goethe published his second novel, Wilhelm Meister's Apprenticeship  ...} \\
        \multicolumn{2}{l}{\textbf{Gold correction}: Johann Wolfgang von Goethe published Wilhelm Meister's Apprenticeship} \\
        \arrayrulecolor{black!20}\midrule
        \textbf{Candidate correction (A)}: Johann Wolfgang von Goethe published Wilhelm Meister's Apprenticeship. & \textbf{DocNLI + ROUGE-1 (A)}: \markred{0.0203}~ + 0.9000\\
        \textbf{Candidate correction (B)}: Johann Wolfgang von Goethe failed to published Wilhelm Meister's Apprenticeship. & \textbf{DocNLI + ROUGE-1 (B)}: 0.1011 + 1.0000 \\
        \multicolumn{2}{l}{\textbf{\modelshort~'s output}: \hlc{lightred}{Johann Wolfgang von Goethe failed to publish Wilhelm meister's Apprenticeship.}} \\

        \arrayrulecolor{black}\bottomrule
    \end{tabular}
    }
    \end{adjustbox}
    \vspace{-2mm}
    \caption{Example outputs from \modelshort~ on the \textsc{FEVER} dataset where it fails to produce faithful factual error corrections. The three types of errors correspond to mistakes in \textit{annotation}, \textit{question answering}, and \textit{correction scoring}.}  
    
    \label{tab:neg_examples}
\end{table*}

\section{Software and Hardware Configurations}
All experiments were conducted on a Ubuntu 18.04.6 Linux machine with a single NVIDIA V100. We use PyTorch 1.11.0 with CUDA 10.2 as the Deep Learning framework and utilize Transformers 4.19.2 to load all pre-trained language models.

\section{Number of Parameters}
The number of parameters for each component in \modelshort~ is provided in parentheses: MixQG-base (220M), UnifiedQA-v2-base (220M), QA-to-claim (220M), DocNLI (355M).

\section{Scientific Artifacts}
The licenses for all the models and software used in this paper are listed below in parentheses: Spacy (MIT License), Stanza (Apache License 2.0), MixQG-base (BSD-3-Clause License), UnifiedQA-v2 (Apache License 2.0), T5-base (Apache License 2.0), DocNLI (BSD-3-Clause License), py-ROUGE (Apache License 2.0), \textsc{FactCC} (BSD-3-Clause License), \textsc{QAFactEval} (BSD-3-Clause License), \textsc{SARI} (GPL-3.0 License), \textsc{BARTScore} (Apache License 2.0).

\end{document}